\pdfoutput=1
\documentclass[10pt, conference, compsocconf]{IEEEtran}
\usepackage{xspace}
\usepackage{gensymb} 

\usepackage{cite}

\ifCLASSINFOpdf
  \usepackage[pdftex]{graphicx}
  \graphicspath{{./img/}}
  \DeclareGraphicsExtensions{.pdf,.jpeg,.png}
\else
\fi
\usepackage[cmex10]{amsmath}
\usepackage[caption=false,font=footnotesize]{subfig}
\usepackage{url}

\makeatletter
\DeclareRobustCommand\onedot{\futurelet\@let@token\@onedot}
\def\@onedot{\ifx\@let@token.\else.\null\fi\xspace}

\def\eg{\emph{e.g}\onedot} 
\def\ie{\emph{i.e}\onedot}

\def\etal{\emph{et al}\onedot}
\makeatother

\makeatletter
\def\ps@IEEEtitlepagestyle{%
  \def\@oddhead{\hfill 2018 15th Conference on Computer and Robot Vision, DOI 10.1109/CRV.2018.00038}%
  \def\@evenhead{}%
  \def\@oddfoot{\mycopyrightnotice}%
  \def\@evenfoot{}%
}
\def\mycopyrightnotice{%
  \begin{minipage}{\textwidth}\ \\%
  {\footnotesize \textcopyright{} 2018 IEEE. Personal use of this material is permitted. Permission from IEEE must be obtained for all other uses, in any current or future media, including reprinting/republishing this material for advertising or promotional purposes, creating new collective works, for resale or redistribution to servers or lists, or reuse of any copyrighted component of this work in other works.}%
  \end{minipage}
  \gdef\mycopyrightnotice{}%
}
\makeatother

\DeclareMathOperator*{\argmax}{argmax}

\begin{document}
%
\title{Do-It-Yourself Single Camera 3D Pointer Input Device}


\author{\IEEEauthorblockN{Bernard Llanos, Yee-Hong Yang}
\IEEEauthorblockA{Department of Computing Science\\
University of Alberta\\
Edmonton, Canada\\
Email: \{llanos, yang\}@cs.ualberta.ca}
}


%


\maketitle

\begin{abstract}
We present a new algorithm for single camera 3D reconstruction, or 3D input for human-computer interfaces, based on precise tracking of an elongated object, such as a pen, having a pattern of colored bands.
To configure the system, the user provides no more than one labelled image of a handmade pointer, measurements of its colored bands, and the camera's pinhole projection matrix.
Other systems are of much higher cost and complexity, requiring combinations of multiple cameras, stereocameras, and pointers with sensors and lights.
Instead of relying on information from multiple devices, we examine our single view more closely, integrating geometric and appearance constraints to robustly track the pointer in the presence of occlusion and distractor objects.
By probing objects of known geometry with the pointer, we demonstrate acceptable accuracy of 3D localization.
\end{abstract}

\begin{IEEEkeywords}
object tracking; pattern recognition; 3D pose estimation; user input devices
\end{IEEEkeywords}

%
\IEEEpeerreviewmaketitle

\section{Introduction}

While computer interfaces have taken over many of the functions of tools such as pointers and pen and paper, they often do not provide the same level of physical intuition.
Fortunately, with digital sensors and computer vision algorithms, we can make computers perceive the physical manipulation of objects, in order to create computer-enabled versions of traditional tools.
In this work, we study computer perception of a pen\slash pointer, and examine its suitability for 3D reconstruction, inspired by the use of slender objects as carving tools.

Many computer interface devices mimic the pen, such as styluses for digital tablets, and devices which can be tracked separately from a screen or writing surface~\cite{Wilson:2003:XUI:642611.642706,Cao:2003:VIT:964696.964716,RefWorks:doc:59d10eefe4b05ddcd6e042ff}.
Others have used such devices for 3D reconstruction of objects by touch.
For objects with deep concavities~\cite{RefWorks:doc:59d10eefe4b05ddcd6e042ff} or transparent or specular surfaces~\cite{RefWorks:doc:58c4d4c7e4b0b605c76b86d3}, 3D reconstruction by probing succeeds where conventional 3D reconstruction techniques fail.
Moreover, remote tracking of a probe is more flexible and cost effective than coordinate measurement machines, which use mechanical sensors to determine 3D locations~\cite{RefWorks:doc:58c4d4c7e4b0b605c76b86d3}.

In parallel to work on high-quality, precise interface devices, others have developed systems which are more accessible because of their low cost and simplicity.
In particular, Chen, Healey, and Amant use color and geometric constraints to track an intuitive 6 DoF input device from a single camera~\cite{Chen:2017:PCC:3141475.3141492}.
Inspired by such low-cost, simple systems, we developed a pen\slash pointer device which is an ordinary pen, or similarly-shaped object, having measured colored bands (\eg colored tape) such that its 3D position can be determined using only a single camera.

Our pointer device presents an interesting object detection challenge, because of distractor colors in the environment, the slender profile of the object, blur, and occlusion.
As such, our pointer detection process, which alternates between applying different appearance and geometric constraints, is our primary contribution, explained in this article.
In the following sections, we begin with a review of previous work (Section \ref{sec:relatedWork}), then explain the algorithm used to obtain the 3D position of the pointer (Section \ref{sec:algorithm}), and evaluate its performance (Section \ref{sec:results}).
We conclude with a summary, and a discussion of the limitations and potential extensions of our work (Section \ref{sec:conclusion}).

\section{Related work}
\label{sec:relatedWork}

\subsection{Devices for pointing and probing}

Most past prototype writing or pointing tools involve complicated hardware setups.
For example, in older virtual reality systems, multiple infrared cameras are used to track retroreflective markers on handheld pens~\cite{RefWorks:doc:58f38c50e4b042050d44849e}. Pointer devices often contain sensors. For instance, both the XWand pointer~\cite{Wilson:2003:XUI:642611.642706} and the SmartPen~\cite{RefWorks:doc:59d10eefe4b05ddcd6e042ff} are equipped with an accelerometer, a magnetometer, and a gyroscope. The former has one infrared LED, whereas the latter has four infrared LEDs, and both are tracked from two views.

Of greater relevance to our work are devices whose poses are estimated using only visual tracking, but here, multiple views are also the norm.
For instance, the VisionWand is a colored rod with differently-colored ends, tracked using two cameras~\cite{Cao:2003:VIT:964696.964716}.
More notably, Michel, Zabulis, and Argyros track a tool, made from a wand attached to a sphere, with four cameras~\cite{RefWorks:doc:58c4d4c7e4b0b605c76b86d3}. The larger number of cameras allows them to perform space carving for 3D reconstruction with an accuracy and precision each around 1 mm.

\subsection{Object detection and tracking}

As a first step in tracking user interface devices, many authors use blob detection methods~\cite{Wilson:2003:XUI:642611.642706, RefWorks:doc:59d10eefe4b05ddcd6e042ff, Chen:2017:PCC:3141475.3141492, RefWorks:doc:58f38c50e4b042050d44849e}. Background subtraction is also very effective~\cite{Chen:2017:PCC:3141475.3141492}, especially in the case of the XWand, which has a flashing infrared LED that can be detected by comparing consecutive images~\cite{Wilson:2003:XUI:642611.642706}. In our work, we avoid background subtraction, so that we can process single images, or images taken by moving cameras (\eg in mobile computing devices).

Once candidate image positions of primitive features of a device (\eg LEDs) are detected, the image positions must be correctly associated with features of the device. Some authors use simple data association methods, such as picking the brightest pixel in the image~\cite{Wilson:2003:XUI:642611.642706}, or applying epipolar constraints from multiple views~\cite{RefWorks:doc:58f38c50e4b042050d44849e}. Others use RANSAC to select a homography between feature detections and known points on the device~\cite{Chen:2017:PCC:3141475.3141492} or directly match length ratios in the image with measurements of the device~\cite{RefWorks:doc:59d10eefe4b05ddcd6e042ff}. Michel, Zabulis, and Argyros use particle swarm optimization to keep track of multiple hypotheses over time and avoid prematurely rejecting candidate positions~\cite{RefWorks:doc:58c4d4c7e4b0b605c76b86d3}.

Looking at object detection in a broader context, we examined solutions for the bin picking problem, in which a robot must select objects from collections, such as within an assembly line. Piccinini, Prati, and Cucchiara developed a reliable method for locating objects using local image feature detection and matching~\cite{PICCININI2012573}, but it is not well-suited for objects with featureless surfaces, and assumes a Euclidean transformation between calibration views of objects and the observed images.

Considering the features of our pointers which could be most reliably extracted from images, we surveyed methods for detecting lines, and colors. Approaches for extracting lines, such as the Hough transform~\cite{Duda:1972:UHT:361237.361242}, or more recent variations and alternatives~\cite{XU20154012,10.1007/978-3-642-13772-3_8} could be used to obtain the midline of a pointer's image, which unfortunately is not the projection of its 3D centerline. Extracted contours, in constrast, can be used to estimate 3D poses, even from a single view, for solids of revolution~\cite{RefWorks:doc:58f38c50e4b042050d44849b, RefWorks:doc:58f38c50e4b042050d448495}.
Unfortunately, these methods would be unstable for our application, as our pointers may be so thin that they are one-dimensional structures in images.

Color detection methods can extract regions of arbitrary shape, not only lines, and can recognize textureless objects that may be missed by local feature detectors.
Works on color detection considered color spaces which could make object recognition invariant to changes in lighting and surface orientation~\cite{RefWorks:doc:58e3d41ee4b0a2720af1aecf, RefWorks:doc:58e3d41de4b0a2720af1aecc}, and introduced techniques for mitigating noise, such as by matching variable kernel density estimators instead of histograms~\cite{RefWorks:doc:58e3d41de4b0a2720af1aecd}.

We closely follow the work of Gevers, Smeulders, and Stokman ~\cite{RefWorks:doc:58e3d41ee4b0a2720af1aecf, RefWorks:doc:58e3d41de4b0a2720af1aecc, RefWorks:doc:58e3d41de4b0a2720af1aecd}, choosing to detect colors using hue variable kernel density estimators. Hue remains constant across a wide variety of viewing and illumination conditions, but its high stability reduces its capacity for discriminating between colors.
Consequently, other authors use two color dimensions for object detection: the two chrominance channels from the YUV color space~\cite{RefWorks:doc:58ec67e1e4b0cc37dc2c9da0}, or both hue and saturation~\cite{Chen:2017:PCC:3141475.3141492}.
These authors also update their color detection models over time, perhaps to compensate for the lower stability of two color dimensions compared to hue alone. Our solution is to use ``one and a half'' color dimensions for detection, and to rely on geometric constraints to overcome shortcomings in color discrimination, as described in Section \ref{sec:detection}.

\section{Algorithm}
\label{sec:algorithm}

\begin{figure}[!t]
\centering
\includegraphics[width=6.2cm]{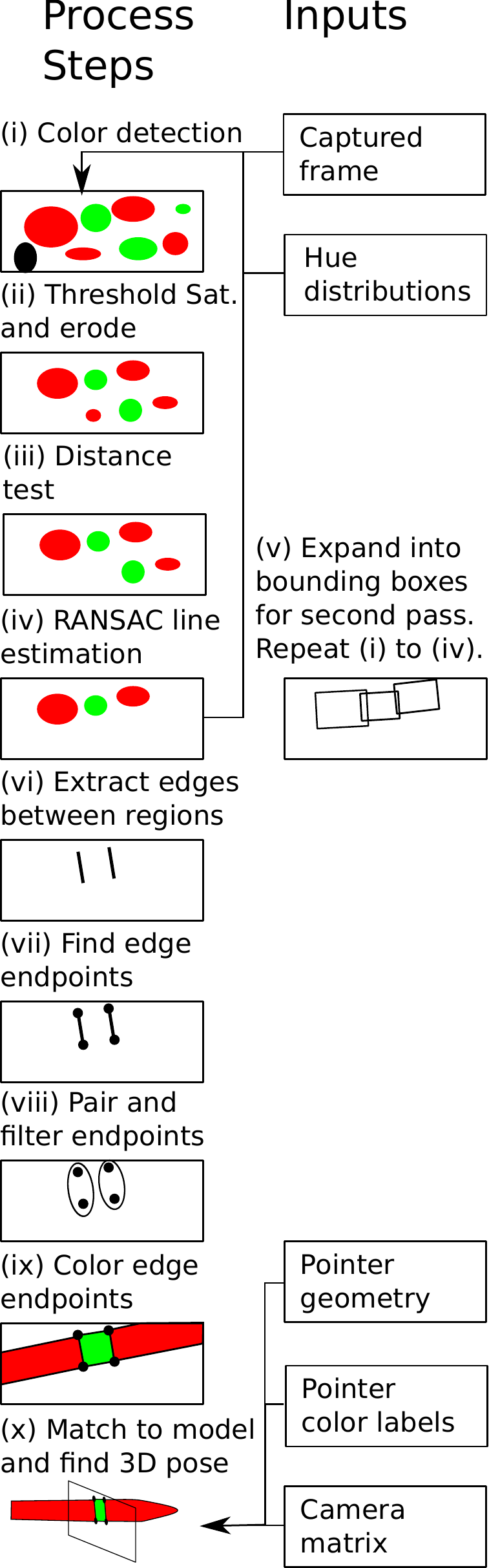}
\caption{An process diagram illustrating the stages in pointer detection (Section \ref{sec:detection}) and pose estimation (Section \ref{sec:poseEstimation}).
The meaning of step (v) is as follows: After the initial colored region detection steps (i) to (iv), bounding boxes are created around the regions (step (v)), and steps (i) to (iv) are repeated within the bounding boxes.}
\label{fig:process}
\end{figure}

Our system obtains 3D positions after three steps: User-assisted color and geometrical calibration for the pointer device (Section \ref{sec:calibration}), automatic detection of the pointer in images from the camera (Section \ref{sec:detection}), and automatic localization of the pointer in 3D space (Section \ref{sec:poseEstimation}).
The automated steps are summarized in Fig.~\ref{fig:process}.
The entire procedure, including building the pointer device, can be completed in under an hour.

\subsection{Pointer design}
\label{sec:design}

In contrast with most 3D input methods, which require specific physical markers or hardware, our system adapts to the pointing device that the user wishes to use.
We therefore make it easy for the user to switch to a different pointer that can perform better on the task at hand.

We require the pointer to be a cylindrically-symmetric object (not necessarily with straight sides), having a pattern of bands along its length.
Each band need not be monochromatic, but should have colors separated in the hue color dimension from the colors of adjacent bands.
In practice, patterns consisting of no more than two or three hues work well, as larger numbers of hues worsen color discrimination.
Non-adjacent bands can have the same colors, and the configuration of bands is arbitrary, so long as the pattern is geometrically or colorimetrically distinct from its reversal, to disambiguate orientation.
Ideally, the pointer should have many bands, as long as each is large enough to be reliably detected at the typical viewing distance, despite blur and noise, and there is a large variance in the lengths of the bands.
Under these two conditions, the system will correctly identify the indices of visible bands in the pattern, even if most of the pattern is occluded.

\begin{figure}[!t]
\centering
\includegraphics[width=8.5cm]{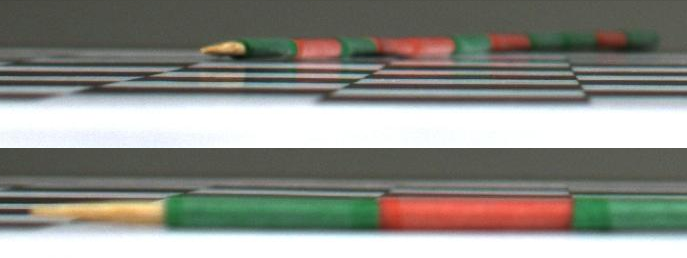}
\caption{This bamboo skewer pointer was reliably detected at camera space $z$-coordinates (\ie depths) from $40$ cm (bottom) to $61$ cm (top), and at angles from approximately $0\degree$ (bottom) to approximately $71\degree$ (top) to the image plane.
The depth of $61$ cm was the maximum tested, because of constraints from the size of the chequerboard used to measure true positions.
Note the large blur in the images, which did not impede pointer detection.}
\label{fig:gridviews}
\end{figure}

In our experiments, we created pointers by wrapping colored tape around bamboo skewers (Fig.~\ref{fig:gridviews}), pens, and a knife sharpening steel.
Although 2D markers, such as ARToolkit markers~\cite{Kato:1999:MTH:857202.858134}, may allow for more accurate and unambiguous 3D positioning than a 1D color pattern, they may be more difficult for the user to fix in a precise location on the pointer, and would make the device less ergonomic.

\subsection{Calibration}
\label{sec:calibration}

For calibration, the user must provide three things: First, an annotated image of the pointer, from which the system constructs one hue variable kernel density estimator for each distinguishable band appearance in the pattern of colored bands, using the method of Gevers and Stokman~\cite{RefWorks:doc:58e3d41de4b0a2720af1aecd}.
Second, the distance from the tip of the pointer, and the pointer's diameter, at each of the edges between colored bands.
Third, lens distortion parameters and the camera projection matrix, such as using well-known algorithms~\cite{RefWorks:doc:5a6651c8e4b0e2d6ab80d4ed, RefWorks:doc:5a6651c8e4b0e2d6ab80d4ee}.


\subsection{Pointer detection}
\label{sec:detection}

Pointer detection provides information about the pointer's image location for accurate determination of the pointer's position in 3D space.
Therefore, the detection process must be invariant to perspective distortion, and must search for features of the pointer that are well-localized.
To satisfy these requirements, we detect the points along the image contour of the pointer that are at junctions between colored bands.
We do not search for the tip of the pointer, as it may be occluded, and because it is only adjacent to at most one detectable colored band.
Points between colored bands are more precisely detected than the pointer's tip because their positions are constrained by two color detections.
We also do not attempt to recover the orientation of the pointer by finding its midline in the image, as the midline is not the image of the pointer's axis of 3D symmetry.

We start, in step (i) in Fig.~\ref{fig:process}, with hue-based color detection, although hue is ill-defined for unsaturated colors~\cite{RefWorks:doc:58e3d41de4b0a2720af1aecd}.
To avoid selecting unsaturated distractor colors, we let the user set two thresholds on the saturation values in the image, $s_1$ and $s_2$, where $s_2 < s_1$.
It is difficult to find an optimal saturation threshold, because saturation depends on shading and highlights~\cite{RefWorks:doc:58e3d41de4b0a2720af1aecc, RefWorks:doc:58e3d41ee4b0a2720af1aecf}.
Chen, Healey, and Amant's solution is to set limits on saturation that optimize color detection in the previous frame~\cite{Chen:2017:PCC:3141475.3141492}. Their system can tolerate excessively low saturation thresholds, however, as their interest points are the centres of blobs, not contour points---Saturation generally decreases near object contours because of shading.

Instead, we use two passes to detect the colored regions of the pointer: In the first pass, we use $s_1$ to reduce false positive detections.
In the second pass, we use $s_2$ for more sensitive detection of the contour of the pointer.
The second pass operates within bounding boxes around the regions output by the first pass.
We create oriented bounding boxes by expanding bounding ellipses for each region by $1.1$ times along their major axes, and $1.5$ times along their minor axes.
Note that we chose the parameter values given throughout this section experimentally, by observing the spatial extent of noise.
We found that the given values were suitable for several different pointers and cameras, so the system is relatively stable in the neighborhoods of these values.

Given the hue variable density estimators representing the color classes of the pointer and the image background, $f_i(\theta)$, where $i$ indexes the color classes, and $\theta$ represents an image pixel's hue value, we assign a pixel to the $k$-th class, where $k = \argmax_i f_i(\theta)$. We represent the image background with a uniform hue distribution.
We do not estimate a hue distribution for the image background, in contrast to Chuang \etal, who use the image background to create ratio distributions~\cite{RefWorks:doc:58e3d41de4b0a2720af1aecb}.
In our experience, there is no advantage to modeling the distribution of colors in the image background: If distractor colors are different from the colors of the pointer, they would be rejected regardless, and if they are similar to the colors of the pointer, a model of the background prevents the colors of the pointer from being detected.

Distractor colors are better handled using geometric constraints:
First, we perform morphological erosion, using a disk structuring element with a radius of $r_1$ and $r_2 < r_1$ on the first and second passes, respectively, to reject very small detected regions. ($r_1$ and $r_2$ are small fractions of the image size, such as $5$ and $2$ pixels, respectively.)
Second, of the remaining regions, we select only those whose borders come within a distance of $w_i = 2r_i + 4,\; i \in \{1,2\}$ pixels of the borders of regions detected for adjacent colors in the known pattern on the pointer.

The final geometric constraint is to use RANSAC (as presented in~\cite{RefWorks:doc:59319199e4b06dee1c4ed3ce}) to fit a line to the centroids of all detected regions (step (iv) in Fig.\ref{fig:process}). During RANSAC iteration, we select the line supported by the largest number of inlier regions, where an inlier region is a region which crosses the line. After RANSAC iteration, we output all regions whose centroids are within $3$ standard deviations of distance from the line, since we observed that the requirement of crossing the line is too restrictive.

With a set of regions corresponding to the colors of the pointer, we can locate the points along the pointer's contour.
To begin, we detect edges between colored bands by creating a binary image $I_{b1}$, where pixels are $1$ if they are within $e = 2r_2 + 1$ pixels from regions detected for both of the colors in any of the pairs of colors which are adjacent on the pointer.
Next, we calculate $\phi$, the angle of the second principal component vector of the $1$-valued pixels with respect to the image $x$-axis, and filter the image with a filter kernel defined by:
\begin{equation}
\begin{split}
& H\left(x,y\right) = \\
&\frac{1}{2 \pi \sigma_d \sigma_a}
\exp\left(-\frac{x^2 + y^2}{2 \sigma_d^2} - \frac{\left(\tan^{-1}\left(\frac{y}{x}\right) - \phi\right)^2}{2 \sigma_a^2}\right)
\end{split}\text{.}
\label{eq:filterkernel}
\end{equation}

We set $\sigma_d = e$ pixels and $\sigma_a = \frac{\pi}{12}$. Convolution with $H$ emphasizes edges perpendicular to axis of the pointer, and reconnects edges which were separated by unsaturated regions in the image (\eg highlights).
We binarize the filtered image using a threshold of $0.3$ to obtain a binary image $I_{b2}$. The image $I_{b3} = I_{b1} \wedge I_{b2}$ is the output represented in step (vi) of Fig.~\ref{fig:process}.

We fit a line, $L_1$, to the $1$-valued pixels in $I_{b3}$. We then select, for each connected component in $I_{b2}$, if such a pair of pixels exists, the two pixels maximally separated in the direction perpendicular to $L_1$ that are also $1$-valued in $I_{b3}$, and that are on opposite sides of $L_1$ (step (vii) of Fig.~\ref{fig:process}). We reject pairs of such pixels where the two pixels are separated by a distance less than $e$. We then look at the coordinates of the pixels in the dimension parallel to $L_1$, and reject pairs of pixels that are not mutual nearest neighbors in this dimension, or that are separated by more than $5$ standard deviations from the mean separation over all pairs.

Finally, we label the pairs of contour points with the colors of the regions on either side (step (ix) of Fig.~\ref{fig:process}). To do so, we clip the detected colored regions to quadrilateral regions bounded by line segments through the detected contour points. The color label on a given side of a pair of contour points is the label of the (clipped) colored region having the closest centroid, in the dimension along $L_2$, to the average coordinate of the two points. $L_2$ is a line fit to the pairs of contour points, and we exclude colored regions which do not cross $L_2$. Consequently, some color labels may be undefined, which represents the image background. An undefined color label does not match with any color during our data association procedure described below.


\subsection{3D pose estimation}
\label{sec:poseEstimation}

\subsubsection{Data association}

Each detected edge between colored bands is described by color labels on either side. Matching only using color labels, a detected edge may be identified with many edges on the pointer, as in the case of the color pattern shown in Fig.~\ref{fig:gridviews}.
We compensate for non-discriminative color labels by aligning the measured and detected color labels using dynamic programming, thus relying on ordering constraints from the 1D geometry of the pointer.
We then use RANSAC~\cite{RefWorks:doc:59319199e4b06dee1c4ed3ce} to find one or more data association hypotheses, selected from the optimal dynamic programming alignment(s), having the highest number of inlier edges.
Each RANSAC hypothesis is a triplet of pairings of detected to known edges, which defines a 1D projective homography between coordinates along the line $L_2$ and measurements of the pointer's colored bands.
The number of inlier edges for a hypothesis is the number of detected coordinates on $L_2$ which, when transformed by the 1D homography, have a (reciprocal) nearest neighbor with consistent color labels.
We select, from the set of hypotheses with the maximal number of inlier edges, the hypothesis which produces the lowest image reprojection error for the final 3D pose estimate.

\subsubsection{Maximum likelihood pose estimation}

The pointer's pose has five degrees of freedom: The 3D coordinates of its tip, $\mathbf{X}_0$ and the two orientation angles representing the unit direction vector, $\mathbf{\hat{d}}$, from its tip to its other end.
As discussed by Zhang, a sequence of collinear points also has five degrees of freedom~\cite{RefWorks:doc:58e3d41de4b0a2720af1aece}, so it is possible to find the pose of an arbitrarily thin pointer.

We estimate the pointer's pose using the pinhole projection equation relating an image position on the pointer's contour, $\mathbf{x}_{ij},\: i\in (1, 2,\ldots, n),\: j\in \{-1, 1\}$ ($j$ distinguishes between the two endpoints of each detected edge between colored bands), with a 3D point, $\mathbf{X}_{ij}$:
\begin{equation}
\mathbf{x}_{ij} = \lambda_{ij} \mathbf{P} \mathbf{X}_{ij}\text{,}
\label{eq:projection}
\end{equation}
where $\mathbf{P}$ is the camera matrix, and $\lambda_{ij}$ is a homogenous scaling factor~\cite{RefWorks:doc:59319199e4b06dee1c4ed3ce}. $\mathbf{X}_i = \mathbf{X}_0 + b_i \mathbf{\hat{d}} + j w_i \mathbf{\hat{u}}$, where $b_i$ is the distance along the axis of the pointer from $\mathbf{X}_0$ to the $i$-th edge. $w_i$ is the radius of the pointer at the $i$-th edge, and $\mathbf{\hat{u}}$ is a unit vector normal to the plane containing the camera center and the axis of the pointer.

Given the detected contour points, $\mathbf{x}^{est}_{ij}$, we use the Levenberg-Marquardt algorithm to minimize the reprojection error, $\sum_{i,j} \lVert\mathbf{x}^{est}_{ij} - \mathbf{x}_{ij}\rVert^2$.
Assuming Gaussian noise in the detected image positions, the result is a maximum likelihood estimate of the pointer's pose.
At least three point correspondences \eqref{eq:projection} are required, because each provides two independent constraints~\cite{RefWorks:doc:59319199e4b06dee1c4ed3ce}.
If the pointer is thin, the image points should correspond to at least three different edges between the colored bands of the pointer, for stability.

We initialize the Levenberg-Marquardt algorithm by solving for the camera-space depths, $v_0$ and $v_n$, of approximations of the points $\mathbf{X}_0$, and $\mathbf{X}_n = \mathbf{X}_0 + b_n \mathbf{\hat{d}}$, respectively.
First, we use the 1D homography from the data association hypothesis to find approximate image projections of $\mathbf{X}_0$, and $\mathbf{X}_n$ along the line $L_2$.
Then we solve a linear system derived from the approximations $\mathbf{x}^{est}_i \approx \lambda_i \mathbf{P} \mathbf{X}_i$, where $\mathbf{x}^{est}_i$ is the average of the projections of $\mathbf{x}^{est}_{ij}$ onto $L_2$. The unknowns are $v_0$, $v_n$, and the $\lambda_i$, such that the solution only determines the ratio of $v_0$ and $v_n$. We correct the scale ambiguity using the known distance between $\mathbf{X}_0$, and $\mathbf{X}_n$, and correct the sign ambiguity by constraining $\mathbf{X}_0$, and $\mathbf{X}_n$ to be in front of the camera.
Our initialization of the Levenberg-Marquardt algorithm is stable even if the detected contour points are collinear.

\section{Experiments and results}
\label{sec:results}

\subsection{Implementation}

To facilitate rapid prototyping, we implemented our algorithm in MATLAB, and made use of many built-in image processing functions.
Aside from the runtime overhead of MATLAB relative to other programming environments, our implementation runs slowly because the built-in functions process irrelevant regions of the image.
Furthermore, we have not explicitly parallelized our code, nor have we leveraged GPU processing, so there is considerable room for improvement before considering simpler, though less precise and robust versions of our algorithm.

Regardless, our implementation can process live video, at a low framerate, because it does not rely on consistency between frames. On a computer with an Intel Core i7 3.6 GHz CPU, it processes images from a webcam, having a resolution of $640 \times 480$ pixels, at $2$-$5$ FPS, and from a Point Grey BlackFly Flea3 camera, having a resolution of $2448 \times 2048$ pixels, at around $0.5$ FPS. We used the BlackFly camera to collect the results described below.

\subsection{Performance characterization}
\label{sec:performance}

\begin{figure*}[!t]
\centerline{%
\subfloat{\includegraphics[width=9cm]{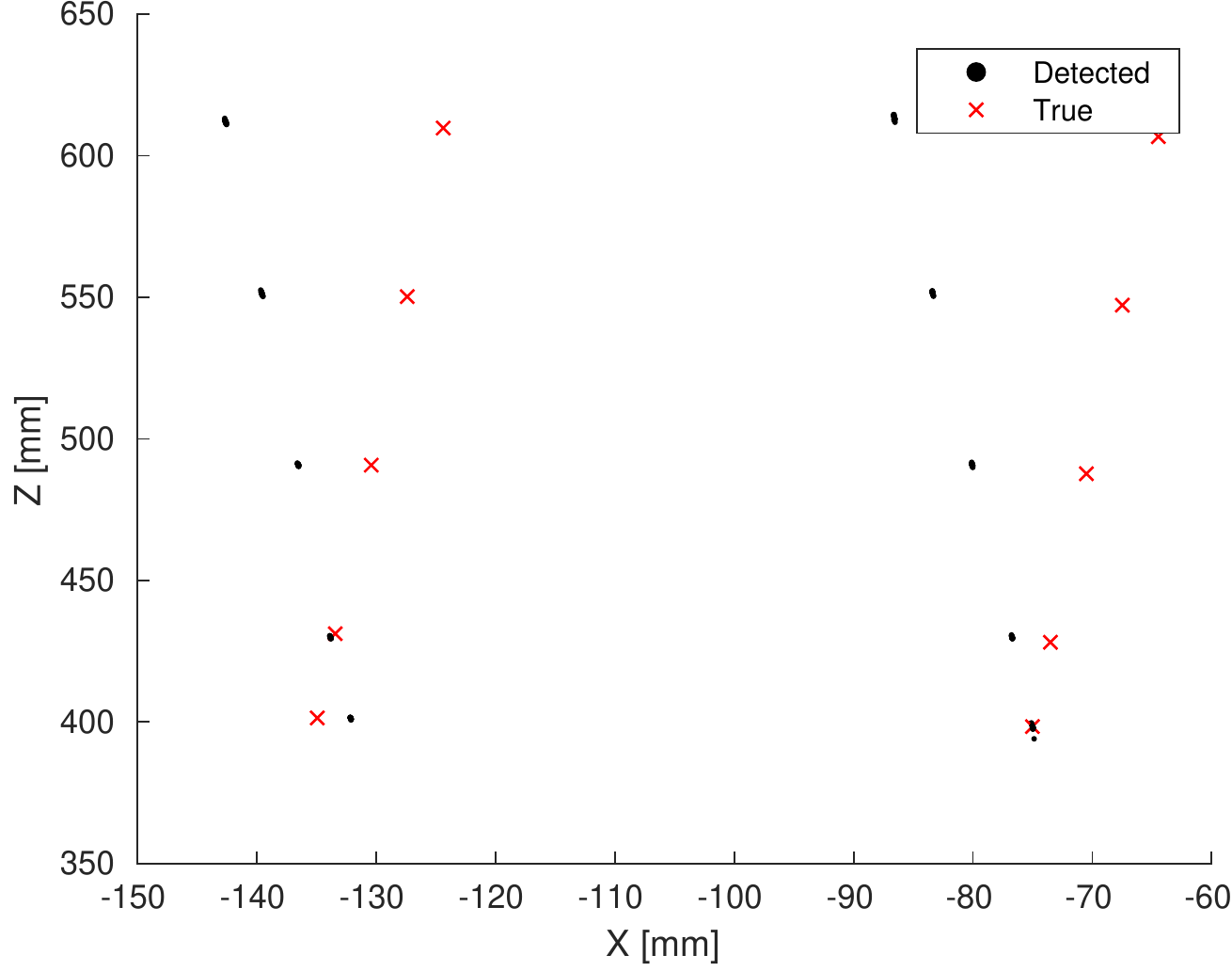}%
\label{fig:angle0points}}
\hfill
\subfloat{\includegraphics[width=9cm]{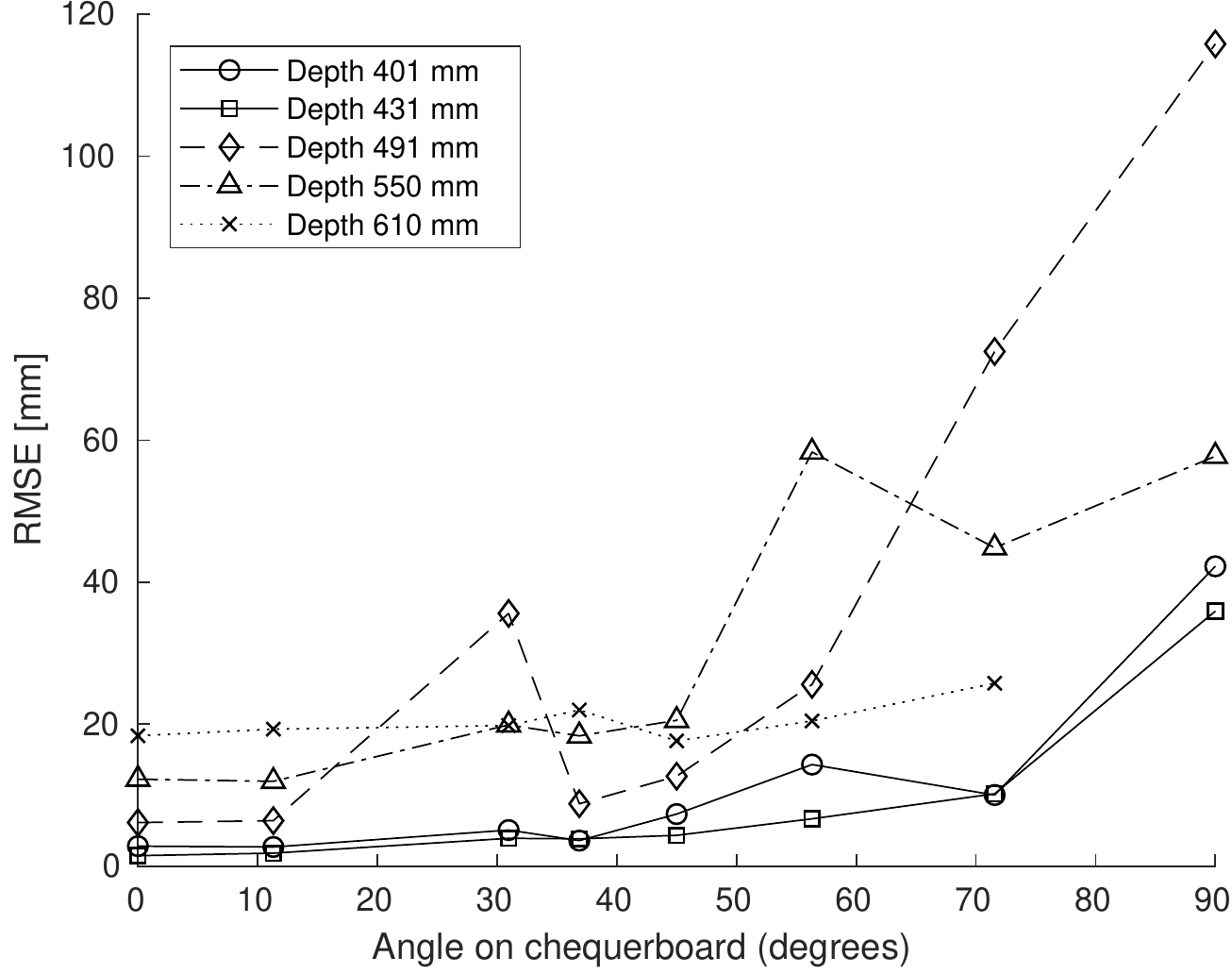}%
\label{fig:gridrmse}}}
\caption{Evaluation of 3D localization by using a chequerboard as a reference object: (Left) Detected vs.~true 3D positions (in the camera's frame of reference) of the pointer's tip, when the pointer was placed at various positions on a chequerboard, and aligned approximately parallel with the image plane (as in the lower half of Fig.~\ref{fig:gridviews}). The detected and true points are registered as follows: First, the true points were rotated parallel to the plane fit to the detected points using principal components analysis. Second, the true points were translated such that the bottom rightmost points in the figure (with the minimum $x$ and minimum $z$-coordinates) correspond. Third, the true points were rotated in the plane such that the vectors from the bottom rightmost to the bottom leftmost points align. (Right) Root-mean-squared distances between the detected and true points, for the line of points on the left side of Fig.~\ref{fig:angle0points}. The datapoint for the most extreme viewing angle and greatest depth is missing because pointer detection failed under these conditions.}
\label{fig_sim}
\end{figure*}

We placed a pointer created from a bamboo skewer (length $251$ mm), and red and green colored tape, at various locations on a chequerboard, and at angles ranging from parallel to the image plane to perpendicular to the image plane.
We did not use the chequerboard for estimating the pose of the camera, nor does our system require it for estimating the pose of the pointer.
Instead, the chequerboard provided hand-measured ground truth positions and orientations of the pointer.
For each of the pointer's poses on the chequerboard, we captured $30$ frames using a Point Grey BlackFly Flea3 camera.
Portions of representative images are provided in Fig.~\ref{fig:gridviews}.
We registered the resulting set of 3D pose estimates to the true poses by anchoring the two together at the most reliably detected points, as described in Fig.~\ref{fig:angle0points}.

The root-mean-squared distances between the true and estimated 3D positions of the pointer's tip are affected primarily by the angle of the pointer to the image plane, as shown in Fig.~\ref{fig:gridrmse}.
The error in pose estimation also increases with the depth of the pointer in the camera's frame of reference, although partly because the detected 3D positions were aligned with the ideal 3D positions at the closest depth, as described in Fig.~\ref{fig:angle0points}.
Part of the effect of depth on the RMS error may result from error in the camera's intrinsic matrix, which could explain why the grid shown in Fig.~\ref{fig:angle0points} is skewed.

The depth of the pointer's tip is estimated with higher uncertainty than any other dimensions of its position.
We computed the principal component directions of the detected points for each chequerboard position and pointer orientation angle, and observed that the first principal component was always aligned with the $z$-axis of the camera's coordinate space.

Intuitively, as the imaging conditions approach affine projection, the pointer's image provides a much weaker constraint on the pointer's depth, because affine imaging is invariant to object depth.
Depth also affects accuracy through the limited resolution of the camera, since depth magnifies the spatial error that results from pixel quantization of the detected point positions.
We would ideally perform further experiments to disentangle the effects of camera field of view angle, pointer depth, and the viewing angle subtended by the pointer. In the current experiment, the camera's field of view angle was fixed, whereas the pointer's depth and the viewing angle subtended by the pointer were varied simultaneously.

The image projections of the second and third principal components of the detected points were relatively more perpendicular and more parallel, respectively, to the image projection of the estimated orientation vector (from tip to tail) of the pointer. We observed that this relationship was true regardless of the angle of the pointer with respect to the image plane.
We conclude that our detection of multiple edges between colored bands allows for accurate localization of the pointer in the dimension parallel to its axis of cylindrical symmetry.
In comparison, our localization of the pointer in the dimension perpendicular to its axis of symmetry is slightly less reliable, most likely because the widths of the detected colored regions are affected by highlights and shadows at the contour of the pointer.
Note that our detector is robust to defocus, as illustrated by the visible defocus in Fig.~\ref{fig:gridviews}.

\subsection{Single view 3D reconstruction}

\begin{figure}[!t]
\centering
\includegraphics[width=8.5cm]{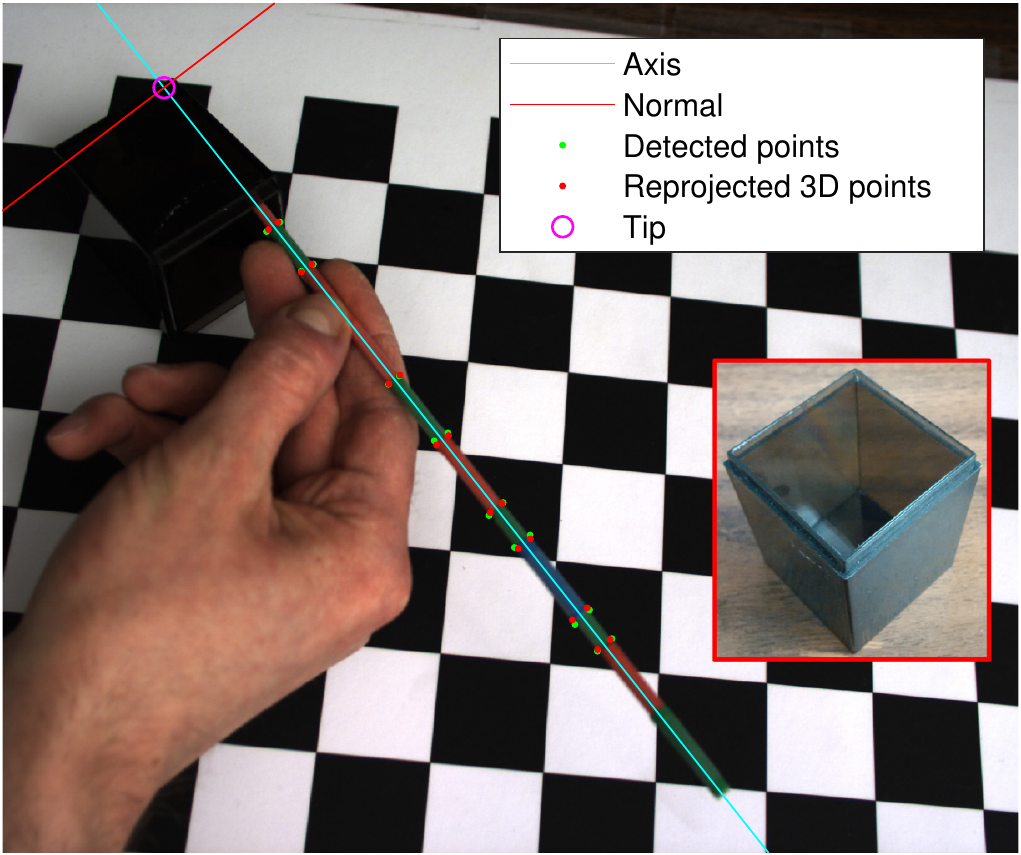}
\caption{Use of a pointer for 3D reconstruction of the bottom of a container, as seen by the Point Grey BlackFly Flea3 camera that collected the input images.
The inset (outlined in red) is provided to give a better idea of the container's shape.
The chequerboard is present only to give a neutral background that can be reliably distinguished from the colors of the pointer.}
\label{fig:squaretracing}
\end{figure}

In Fig.~\ref{fig:squaretracing}, we show the experimental setup for 3D reconstruction of the bottom of a container used to hold paper clips.
The container is made of dark, textureless, reflective plastic, so its interior would be difficult to reconstruct using vision alone, even without occlusion by the container's walls.
In this setting, our system is challenged, by occlusion of the first two edges between colored bands by the container walls, and by the hand, which has a similar hue to the red bands of the pointer, and which effectively hides an additional two edges (as seen in Fig.~\ref{fig:squaretracing}).

With patterns of colored bands containing repeated colors, sometimes the system will not associate detected points with the correct colored bands of the pointer.
In this experiment, we added a single band of blue tape to improve data association, producing a sequence of $10$ edges between colored bands along the length of the pointer, $6$-$8$ of which were usually visible to the camera.
Otherwise, we used the same pointer as in section~\ref{sec:performance} (where it had only red and green bands).
Still, there were occasional data association failures, particularly when there were false positive edge detections (excess, spurious detections).
When data association fails, the estimated position of the pointer's tip is often grossly misestimated.
Conseqeuntly, such points may not seriously affect the reconstruction, if the point cloud is filtered to remove outliers.

\begin{figure}[!t]
\centering
\includegraphics[width=8.5cm]{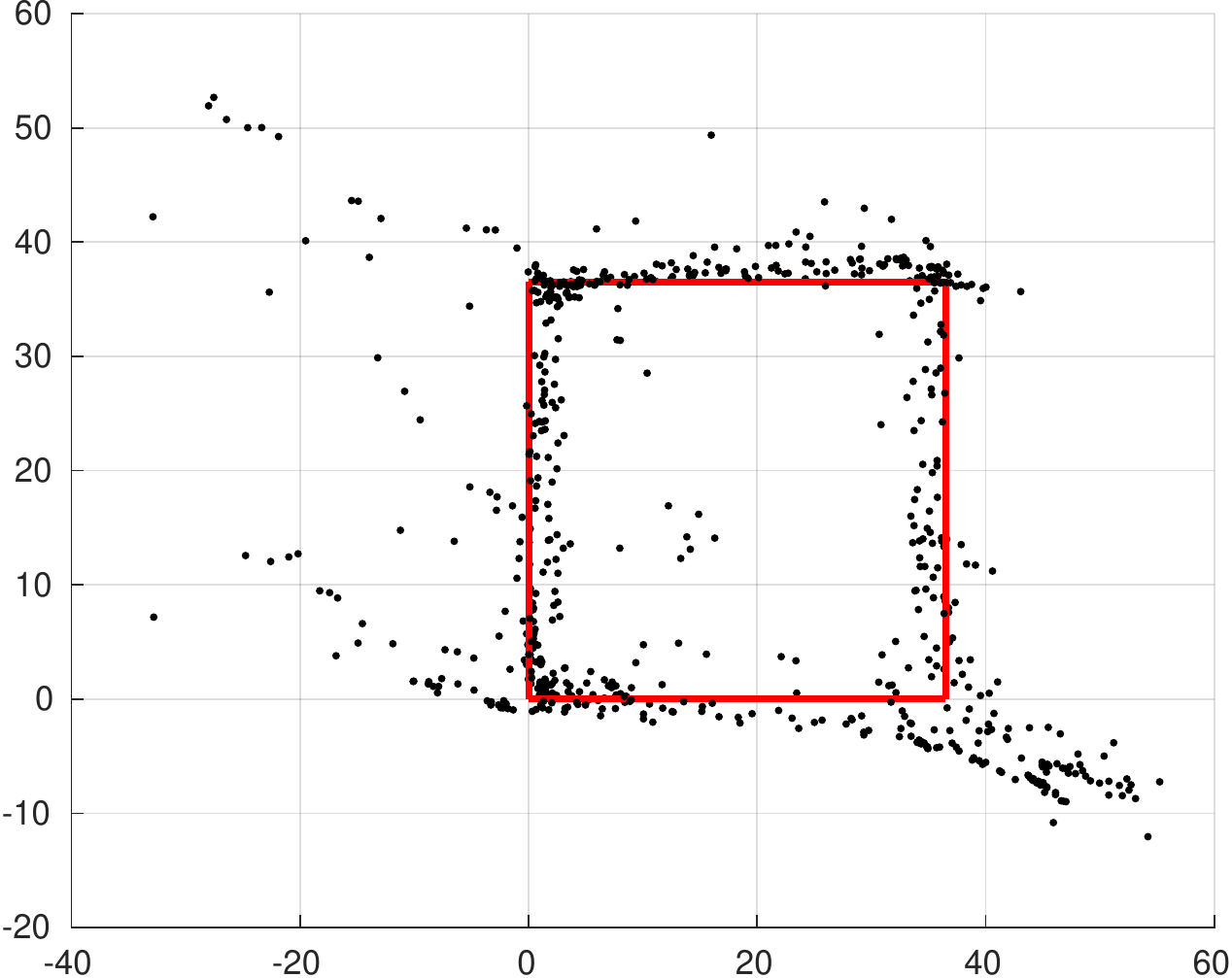}
\caption{Reconstruction (in millimeters) of the bottom of the container shown in Fig.~\ref{fig:squaretracing}. The reconstructed points have been projected to the 2D space in which they are best aligned with the true $37$ mm square (drawn in red). Note that an ideal point cloud would be inside the square because the pointer's tip has a non-negligible diameter of close to 1 mm.}
\label{fig:square}
\end{figure}

A point cloud collected by tracing the bottom of the paper clip container is shown in Fig.~\ref{fig:square}. The bottom of the container is a square of side length $37$ mm, at a distance of approximately $48$ cm from the camera's center of projection.
Extremely poorly estimated points, located outside the region shown in Fig.~\ref{fig:square}, represent $7.1\,\%$ of the data.
$90\,\%$ of all points are within an error (in 3D space) of $21.2$ mm, whereas $50\,\%$ are within an error of $3.4$ mm.
On one hand, the accuracy of the reconstruction is lower than what can be achieved with more complicated and expensive apparatuses (\eg~\cite{RefWorks:doc:58c4d4c7e4b0b605c76b86d3, RefWorks:doc:59d10eefe4b05ddcd6e042ff}).
On the other hand, the accuracy of our system could be improved by using a higher-resolution camera and a longer pointer with more distinctive colors, whereas the accuracy of most other probing techniques could not be improved without time-consuming modifications to their devices or algorithms.

\section{Conclusion}
\label{sec:conclusion}

We proposed a new 3D pointing and probing system that can be set up quickly and without special-purpose materials and devices.
Our system is extremely low cost, yet it is also flexible, as the user can design their own pointing devices.
We do not sacrifice simplicity for flexibility---The user need only provide a single image and measurements of the pointer's color band pattern to calibrate our detector.
Moreover, pointer design is intuitive, because we use the hue and saturation color dimensions for pointer color detection.
Consequently, we enable users to easily troubleshoot challenges in the environment, such as distractor objects.

We show that our system performs well even in the presence of occlusion and defocus, but that its performance declines as the angle between the pointer and the image plane increases.
In general, its accuracy would be suitable for human-computer interaction tasks such as gesture recognition.
Our technique also shows promise for single-view 3D reconstruction, in particular for challenging scenarios, such as probing the interiors of concave objects.
We hope to further characterize how our system behaves across different pointer design choices and camera parameters (\eg field of view size), using motion capture technology to obtain highly accurate ground truth for evaluation.

There are areas we can explore to improve our system.
Most notably, we could rely on temporal consistency between frames to filter out noise in the pointer's estimated pose.
Given temporally-consistent frames, we could integrate object tracking algorithms, such as registration-based object tracking, which could run very quickly~\cite{siam2017}. We could also segment out the user's hand using methods such as that of Zhao, Luo and Quan~\cite{zhao2017}, removing a potential distractor object, and providing useful information for human-computer interaction applications.

Another interesting extension pertains to usability: We could have the user calibrate the camera using their pointer, as described by Zhang~\cite{RefWorks:doc:58e3d41de4b0a2720af1aece}.
As such, the user would not need to construct a separate calibration target.
Presently, our implementation requires camera calibration data as input, and so is usable by those generally familiar with 3D computer vision.
As such, we encourage others in the community to experiment with our prototype implementation\footnote{\url{https://github.com/bllanos/linear-probe}} and share in the ``do-it-yourself'' philosophy that inspired us.


\section*{Acknowledgments}

We thank NSERC, Alberta Innovates, and the University of Alberta for financial support.
We also thank Kevin Gordon, Dale Schuurmans, and Noah Weninger for fruitful conversations, and the anonymous reviewers for their constructive comments.



\bibliographystyle{./IEEEtranBST/IEEEtran}
\bibliography{./IEEEtranBST/IEEEabrv,refs}

\begin{thebibliography}{10}
\providecommand{\url}[1]{#1}
\csname url@samestyle\endcsname
\providecommand{\newblock}{\relax}
\providecommand{\bibinfo}[2]{#2}
\providecommand{\BIBentrySTDinterwordspacing}{\spaceskip=0pt\relax}
\providecommand{\BIBentryALTinterwordstretchfactor}{4}
\providecommand{\BIBentryALTinterwordspacing}{\spaceskip=\fontdimen2\font plus
\BIBentryALTinterwordstretchfactor\fontdimen3\font minus
  \fontdimen4\font\relax}
\providecommand{\BIBforeignlanguage}[2]{{%
\expandafter\ifx\csname l@#1\endcsname\relax
\typeout{** WARNING: IEEEtran.bst: No hyphenation pattern has been}%
\typeout{** loaded for the language `#1'. Using the pattern for}%
\typeout{** the default language instead.}%
\else
\language=\csname l@#1\endcsname
\fi
#2}}
\providecommand{\BIBdecl}{\relax}
\BIBdecl

\bibitem{Wilson:2003:XUI:642611.642706}
A.~Wilson and S.~Shafer, ``{XWand}: {UI} for intelligent spaces,'' in
  \emph{Proceedings of the SIGCHI Conference on Human Factors in Computing
  Systems}, ser. CHI '03.\hskip 1em plus 0.5em minus 0.4em\relax New York, NY,
  USA: ACM, 2003, pp. 545--552.

\bibitem{Cao:2003:VIT:964696.964716}
X.~Cao and R.~Balakrishnan, ``{VisionWand}: Interaction techniques for large
  displays using a passive wand tracked in {3D},'' in \emph{Proceedings of the
  16th Annual ACM Symposium on User Interface Software and Technology}, ser.
  UIST '03.\hskip 1em plus 0.5em minus 0.4em\relax New York, NY, USA: ACM,
  2003, pp. 173--182.

\bibitem{RefWorks:doc:59d10eefe4b05ddcd6e042ff}
B.~Milosevic, F.~Bertini, E.~Farella, and S.~Morigi,
  ``\BIBforeignlanguage{English}{A {SmartPen} for {3D} interaction and
  sketch-based surface modeling},''
  \emph{\BIBforeignlanguage{English}{International Journal of Advanced
  Manufacturing Technology}}, vol.~84, no. 5-8, pp. 1625--1645, 2016.

\bibitem{RefWorks:doc:58c4d4c7e4b0b605c76b86d3}
D.~Michel, X.~Zabulis, and A.~A. Argyros, ``\BIBforeignlanguage{English}{Shape
  from interaction},'' \emph{\BIBforeignlanguage{English}{Machine Vision and
  Applications}}, vol.~25, no.~4, pp. 1077--1087, 2014.

\bibitem{Chen:2017:PCC:3141475.3141492}
Z.~Chen, C.~G. Healey, and R.~S. Amant, ``Performance characteristics of a
  camera-based tangible input device for manipulation of {3D} information,'' in
  \emph{Proceedings of the 43rd Graphics Interface Conference}, ser. GI
  '17.\hskip 1em plus 0.5em minus 0.4em\relax Canadian Human-Computer
  Communications Society, 2017, pp. 74--81.

\bibitem{RefWorks:doc:58f38c50e4b042050d44849e}
M.~Ribo, A.~Pinz, and A.~L. Fuhrmann, ``\BIBforeignlanguage{English}{A new
  optical tracking system for virtual and augmented reality applications},''
  \emph{\BIBforeignlanguage{English}{Conference Record - IEEE Instrumentation
  and Measurement Technology Conference}}, vol.~3, pp. 1932--1936, 2001.

\bibitem{PICCININI2012573}
P.~Piccinini, A.~Prati, and R.~Cucchiara, ``Real-time object detection and
  localization with {SIFT}-based clustering,'' \emph{Image and Vision
  Computing}, vol.~30, no.~8, pp. 573 -- 587, 2012, special Section: Opinion
  Papers.

\bibitem{Duda:1972:UHT:361237.361242}
R.~O. Duda and P.~E. Hart, ``Use of the {Hough} transformation to detect lines
  and curves in pictures,'' \emph{Commun. ACM}, vol.~15, no.~1, pp. 11--15,
  Jan. 1972.

\bibitem{XU20154012}
Z.~Xu, B.-S. Shin, and R.~Klette, ``Closed form line-segment extraction using
  the {Hough} transform,'' \emph{Pattern Recognition}, vol.~48, no.~12, pp.
  4012 -- 4023, 2015.

\bibitem{10.1007/978-3-642-13772-3_8}
M.~Alem{\'a}n-Flores, L.~Alvarez, P.~Henr{\'i}quez, and L.~Mazorra,
  ``Morphological thick line center detection,'' in \emph{Image Analysis and
  Recognition}, A.~Campilho and M.~Kamel, Eds.\hskip 1em plus 0.5em minus
  0.4em\relax Berlin, Heidelberg: Springer Berlin Heidelberg, 2010, pp. 71--80.

\bibitem{RefWorks:doc:58f38c50e4b042050d44849b}
C.~Colombo, A.~D. Bimbo, and F.~Pernici, ``\BIBforeignlanguage{English}{Metric
  {3D} reconstruction and texture acquisition of surfaces of revolution from a
  single uncalibrated view},'' \emph{\BIBforeignlanguage{English}{IEEE
  Transactions on Pattern Analysis and Machine Intelligence}}, vol.~27, no.~1,
  pp. 99--114, 2005.

\bibitem{RefWorks:doc:58f38c50e4b042050d448495}
C.~J. Phillips and K.~Danillidis, ``\BIBforeignlanguage{English}{Absolute pose
  and structure from motion for surfaces of revolution: Minimal problems using
  apparent contours},'' in \emph{\BIBforeignlanguage{English}{4th International
  Conference on 3D Vision, 3DV 2016}}.\hskip 1em plus 0.5em minus 0.4em\relax
  Institute of Electrical and Electronics Engineers Inc., October 2016, pp.
  221--229.

\bibitem{RefWorks:doc:58e3d41ee4b0a2720af1aecf}
T.~Gevers and A.~W.~M. Smeulders, ``\BIBforeignlanguage{English}{Color-based
  object recognition},'' \emph{\BIBforeignlanguage{English}{Pattern
  Recognition}}, vol.~32, no.~3, pp. 453--464, 1999.

\bibitem{RefWorks:doc:58e3d41de4b0a2720af1aecc}
K.~V.~D. Sande, T.~Gevers, and C.~Snoek,
  ``\BIBforeignlanguage{English}{Evaluating color descriptors for object and
  scene recognition},'' \emph{\BIBforeignlanguage{English}{IEEE Transactions on
  Pattern Analysis and Machine Intelligence}}, vol.~32, no.~9, pp. 1582--1596,
  2010.

\bibitem{RefWorks:doc:58e3d41de4b0a2720af1aecd}
T.~Gevers and H.~Stokman, ``\BIBforeignlanguage{English}{Robust histogram
  construction from color invariants for object recognition},''
  \emph{\BIBforeignlanguage{English}{IEEE Transactions on Pattern Analysis and
  Machine Intelligence}}, vol.~26, no.~1, pp. 113--118, 2004.

\bibitem{RefWorks:doc:58ec67e1e4b0cc37dc2c9da0}
A.~A. Argyros and M.~I.~A. Lourakis, ``Real-time tracking of multiple
  skin-colored objects with a possibly moving camera,'' in \emph{8th European
  Conference on Computer Vision}.\hskip 1em plus 0.5em minus 0.4em\relax
  Berlin, Heidelberg: Springer Berlin Heidelberg, May 2004, p. 368.

\bibitem{Kato:1999:MTH:857202.858134}
H.~Kato and M.~Billinghurst, ``Marker tracking and {HMD} calibration for a
  video-based augmented reality conferencing system,'' in \emph{Proceedings of
  the 2Nd IEEE and ACM International Workshop on Augmented Reality}, ser. IWAR
  '99.\hskip 1em plus 0.5em minus 0.4em\relax Washington, DC, USA: IEEE
  Computer Society, 1999, pp. 85--.

\bibitem{RefWorks:doc:5a6651c8e4b0e2d6ab80d4ed}
Z.~Zhang, ``A flexible new technique for camera calibration,'' \emph{IEEE
  Transactions on Pattern Analysis and Machine Intelligence}, vol.~22, no.~11,
  pp. 1330--1334, 2000.

\bibitem{RefWorks:doc:5a6651c8e4b0e2d6ab80d4ee}
J.~Heikkil\"{a} and O.~Silven, ``Four-step camera calibration procedure with
  implicit image correction,'' in \emph{Proceedings of the 1997 IEEE Computer
  Society Conference on Computer Vision and Pattern Recognition}, Anon,
  Ed.\hskip 1em plus 0.5em minus 0.4em\relax Los Alamitos, CA, United States:
  IEEE, June 1997, pp. 1106--1112.

\bibitem{RefWorks:doc:58e3d41de4b0a2720af1aecb}
M.-C. Chuang, J.-N. Hwang, K.~Williams, and R.~Towler, ``Tracking live fish
  from low-contrast and low-frame-rate stereo videos,'' \emph{IEEE Transactions
  on Circuits and Systems for Video Technology}, vol.~25, no.~1, pp. 167--179,
  2015.

\bibitem{RefWorks:doc:59319199e4b06dee1c4ed3ce}
R.~Hartley and A.~Zisserman, \emph{Multiple view geometry in computer vision},
  2nd~ed.\hskip 1em plus 0.5em minus 0.4em\relax Cambridge [u.a.]: Cambridge
  Univ. Press, 2004.

\bibitem{RefWorks:doc:58e3d41de4b0a2720af1aece}
Z.~Zhang, \emph{Camera calibration with one-dimensional objects}, ser. 7th
  European Conference on Computer Vision, ECCV 2002, M.~Nielsen, A.~Heyden,
  G.~Sparr, and P.~Johansen, Eds.\hskip 1em plus 0.5em minus 0.4em\relax
  Springer Verlag, 2002, vol. 2353.

\bibitem{siam2017}
M.~Siam, A.~Singh, C.~Perez, and M.~Jagersand, ``{4-DoF} tracking for robot
  fine manipulation tasks,'' in \emph{2017 14th Conference on Computer and
  Robot Vision (CRV)}, 2017, pp. 329--336.

\bibitem{zhao2017}
Y.~Zhao, Z.~Luo, and C.~Quan, ``Unsupervised online learning for fine-grained
  hand segmentation in egocentric video,'' in \emph{2017 14th Conference on
  Computer and Robot Vision (CRV)}, 2017, pp. 248--255.

\end{thebibliography}

\end{document}